# ConvoWaste: An Automatic Waste Segregation Machine Using Deep Learning

Md. Shahariar Nafiz
Department of Electrical and Electronic Engineering
American International University-Bangladesh (AIUB)
shahariarnafiz48@gmail.com

Shuvra Smaran Das
Department of Computer Science
American International University-Bangladesh (AIUB)
shuvradas59@gmail.com

Md. Kishor Morol
Department of Computer Science
American International University-Bangladesh (AIUB)
kishor@aiub.edu

Abdullah Al Juabir
Department of Computer Science
American International University-Bangladesh (AIUB)
abdullah@aiub.edu

Dip Nandi
Department of Computer Science
American International University-Bangladesh (AIUB)
dip.nandi@aiub.edu

*Abstract*—Nowadays, proper urban waste management is one the biggest concerns for maintaining a green and clean environment. An automatic waste segregation system can be a viable solution to improve the sustainability of the country and to boost up the circular economy. This paper proposes a machine to segregate the waste into the different parts with the help of smart object detection algorithm using ConvoWaste in the field of Deep Convolutional Neural Network (DCNN), and image processing technique. In this paper, the deep learning and image processing techniques are applied to classify the waste precisely and the detected waste is placed inside the corresponding bins with the help of a servo motor-based system. This machine has the provision to notify the responsible authority regarding the waste level of the bins and the time to trash out the bins filled with garbage by using the ultrasonic sensors placed in each bin and the dual-band GSM-based communication technology. The entire system is controlled remotely through an android app in order to dump the separated waste in a desired place by its automation properties. The use of this system can aid the process of recycling resources that were initially destined to become waste, utilizing natural resources and turning these resources back into the usable products. Thus, the system helps to fulfill the criteria of circular economy through the resource optimization and extraction. Finally, the system is made to provide the services at a low cost with higher accuracy level in terms of the technological advancement in the field of Artificial Intelligence (AI). We have got 98% accuracy for our ConvoWaste deep learning model.

## I. INTRODUCTION

Urban Waste management refers to the processes and actions, which are required to manage the waste from its inception to its final disposal in order to ensure a clean environment and proper resource utilization. In every single day, waste is being generated in many ways such as industrial activities, household activities, medical activities and so on. Managing these huge amounts of generated waste by manual operation is very difficult, time consuming and costly. That is why this is the peak time to consider an automatic waste management system in order to recycle and reuse the waste as well as to reduce its production. According to [1], in the European region, 56% or 423 million tons of the generated waste was recycled domestically in 2016 and 24% or 179 million tons of generated waste was gathered in the landfill at the same time. This paper shows that the proper waste management system of the country is one of the most important steps for the waste recycling process. That is why the use of modern technology combined with a proper waste management system can be beneficial and the results of this system would be innumerable if it is implemented properly.

[2] Mentions that the circular economic system has been aiming to increase the use of resources in an efficient way to improve the overall economy and environmental sustainability of a country. The main concept of circular economy (CE) is found in [3] and can be defined as the practice to manage resource circularity, efficiency and optimization that proposes to use waste as resources to create economic value. CE includes both natural and human resources. Therefore, for maintaining CE, the proper urban waste management is one of the most challenging constraints right now for the developing and undeveloped countries as suggested by United Nations (UN) [4].

In Bangladesh, considering the Dhaka city, which is the capital of Bangladesh, 3500 tons of mixing solid waste is being produced everyday [5]. Unfortunately, only 1800 tons out of this huge amount of generated solid waste, is collected and dumped properly following the conventional method of using labor forces and rest of the wastes remain uncollected or take place in the backyards or landfills [5]. With reference to the problems regarding the waste management system and environmental sustainability, some of the waste segregation machines were introduced to prevent the waste pollution caused by mixing waste of different types. Initially, those machines were inaugurated based on the different types of sensors only to classify the waste such as dry waste, wet waste and metal waste [6]. However, the proliferation of urbanization and socialization has changed the trend of using sensor-based application for managing the colossal amounts of generated waste. On the contrary, the technological advancement in AI has created a great impact on recent waste segregation machines. Considering the huge progression in AI, the system has been proposed using ConvoWaste in the field of DCNN, and image processing through this paper can classify six types of waste such as plastic waste, metal waste, glass waste, organic waste, medical waste and e-waste without any human intervention.

A new architecture of image classification has been devised through the application of Capsule-Net. [7] Reflects the waste segregation system for two types of waste such as plastic waste and non-plastic waste using Capsule-Net. However, in spite of extensive development of this model, it has some disadvantages. One of the major disadvantages of this model is the training image dataset. It requires huge training image dataset to achieve its final prediction level for image classification purpose. Conversely, ConvoWaste requires less training image dataset to achieve its ultimate prediction level while comparing to DCNN. Likewise, it retains the hierarchical pose relationship between object parts, and can identify the 3D space.There are some advanced techniques for smart waste management system based on Internet of Things (IoT) and Deep learning (DL). [8] Demonstrates a smart waste segregation system using LoRaWAN communication protocol and Tensorlow based DL model. Moreover, the author has claimed that the coverage area of this communication medium is up to 22 kilometers using dual frequency band of 902-928 MHz. It alludes that for the waste monitoring purpose, the waste segregation machine and the monitoring room or person should be within that region, which is not ideal for the largest cities or communities. Considering this problem, a system has been introduced through this paper based on dual-band Global System for Mobile Communications (GSM) module to get the seamless connectivity between waste segregation machine and monitoring room or person without considering the distance. Also, [9] describes that the dual-band GSM module has much higher bandwidth and the data transfer rate is much faster than the LoRaWAN communication protocol. Considering this drawback, the system is proposed through this paper can aid the dumping process placing the automatic waste segregation machine in a desired place and this feature of this automatic waste segregation machine is fully controlled via an android app, based on the Bluetooth communication medium.

The research objectives of our paper are-

a.  Make a relationship between hardware and software to classify a waste data image.

b.  Propose a noble model and train the model with six different collected waste image datasets.

c.  Get a state of art accuracy and classify an unseen image in real time.

d.  Finally, send the classified data to the hardware system, and according to the data classification the classified garbage is placed into the corresponding bin.

Moreover, rest of the sections are organized as follows. Section II discusses about the related work. Section III provides in details information regarding the architecture and design of the system. Section IV reflects the experimental setup for object detection. Section V describes the hardware design and implementation of the system. Section VI provides the result analysis. Finally, the conclusion part is placed at section VII.

## II.  Related Work

This section provides an overview of the previous works and contributions of past researchers who developed similar approaches to the proposed waste segregation system of this paper. Automatic waste segregation system is not a new term. At the very beginning, the automatic waste segregation system was introduced based on the sensor-based applications [10]. However, the trend of using sensors for waste classification has been changed due to the dynamic innovation in the field of AI. In [11], the system was proposed based on Capsule Neural Network (Capsule-Net) to classify the plastic and non-plastic waste. The authors created two public datasets and found out the accuracy level up to 96.3% and 95.7%. The entire system was developed and tested on several hardware devices. In the study of [12], the waste classification model was proposed based on pre-trained Convolutional Neural Network, which was trained using ResNet-50 and Support Vector Machine (SVM). The accuracy level of that proposed system was 87% that was tested based on the public dataset. As per [13], the authors investigated the classification process of electronics-waste, which was known as the e-waste. The proposed model was based on the CNN and RCNN to classify several types of the e-waste. The authors had enhanced the accuracy level of that proposed system from 90 to 97% but the model was only designed for e-waste classification.

According to paper [14], the author claimed a different waste classification model to classify different types of waste using deep learning algorithm, which was mainly applicable to recycle the garbage. Through the proposed scheme of that paper, a deep learning paradigm was reflected to classify the garbage autonomously. During the thorough investigation of [16], it was found that the authors had developed a cost-effective intelligent trash bin for the waste management purpose. To implement a notification process of the garbage level inside the bins, a GSM-based communication medium was proposed. The overall performance of the entire system was satisfactory, which was claimed by the authors. The IoT based smart waste management system in [17] was implemented to reduce the food waste. The implemented model was integrated with a router and a cloud server to collect and analyze the information regarding the food waste. The experiment was tested successfully in several steps and the amount of food waste was abated by 33%. The waste management robot introduced in [18], used a deep learning and fuzzy logic interference, but the authors did not propose the notification process and the smart dumping system clearly. The authors proposed a system that can segregate recyclable materials with the help of smart bin in combination of three types of the sensors such as inductive sensor, capacitive sensor and photoelectric sensor [19]. From the thorough investigation of the literature review, it is found that no technique was proposed in a single system that can segregate the waste and monitor the entire process remotely using the communication technology. While every model used deep learning and image processing technique to classify and identify the waste properly, there was no clear idea about smart dumping system for the classified waste through the proposed system.

## III.  Architecture and Design of the System

An automatic waste segregation machine is designed and implemented using both sensors and deep learning methods. Initially, a waste collection inlet is used to collect the waste and a conveyor belt is used to convey the waste from one end to another end. Afterwards, the presence of the waste is detected by using an ultrasonic sensor. After detecting the presence of the waste, the conveyor belt creates a delay for 10 seconds to capture the image of the waste using the camera module mounted in the conveyor belt. Then the captured image of the waste is sent to the processing unit (Computer) to process the real-time image of the waste using deep learning method. To classify the waste such as plastic, metal, glass, medical, organic and e-waste, ConvoWaste is used. For deep learning, ConvoWaste is a pre-trained convolutional neural network. Basically, it is trained over our collected dataset. A serial communication is established between an ATmega 2560 AVR microcontroller and the processing unit of the system to send the detected output to microcontroller from the processing unit. When an object is detected, a signal is sent to microcontroller via the serial communication from the processing unit to control the servomotors. The depicted block diagram in Fig.1 shows the step-by-step mechanism of the automatic waste segregation machine.

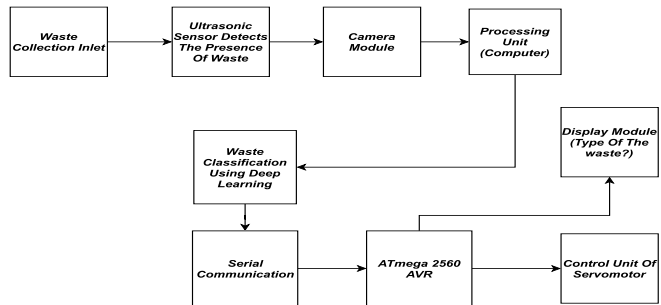

Fig. 1. *Step by step block diagram of the automatic waste segregation machine.*

### A. Flowchart of the Segregation Process

The inserted flowchart in Fig. 2 demonstrates the waste sorting maneuver and the mechanism of the servomotor controller unit of the proposed automatic waste segregation machine.

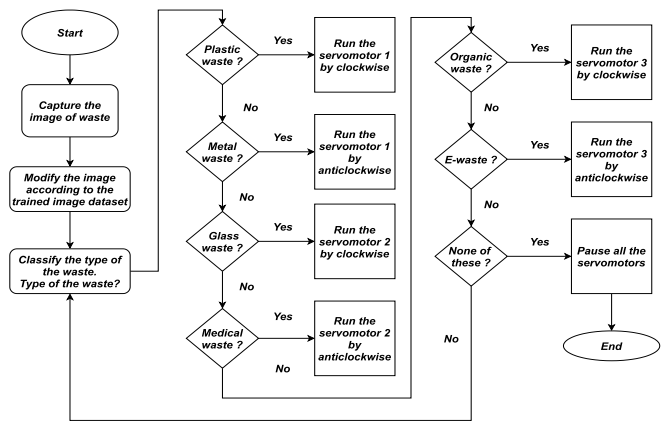

Fig. 2. *Flowchart of the segregation process.*

Initially, the image is captured by using the camera module mounted in the conveyor belt. After capturing the image, the real-time image of the waste is modified according to ConvoWaste and the processing unit of the system displays the type of the processed image precisely. To place the detected waste inside the bins a servomotor-based controller unit is used. Three servomotors are used to place six types of the waste using two directional movement of each servomotor. For instance, when the waste is detected as plastic, then the servomotor 1 is run clockwise to place the waste inside the corresponding bin. Similarly, when the waste is classified as metal, the same servomotor is run anticlockwise to place the waste inside the corresponding bin. The servomotor-based controller unit of this automatic waste segregation machine is controlled via the serial communication between a microcontroller and the processing unit. The above flowchart indicates that if the waste is not detected as any of the types, then all the servomotors are stopped. Otherwise, it starts to follow the same segregation process repeatedly.

Eventually, it is articulated from the above flowchart that the system follows the same process for each step to classify six types of waste.

### B. Transfer Learning With ConvoWaste

For deep learning, transfer learning is one of the most important techniques whereby a neural network is trained on a specific problem similar to the problem that is being solved [20]. There are several layers in the trained neural network but one or more layers are then used in the new model trained on the problem of interest [21]. There are two straightforward processes for transfer learning [22] and these are given below:

- Using the custom image dataset, training of the model is accomplished.

- The initialization of the training model with weights from a pre-trained model.

### C. Overview of ConvoWaste

The ConvoWaste model is used in the experiment. To attain the best level of accuracy, this CNN model combines a pre-trained Inception-Resnet V2 model [15] with several extra layers. To extract features, the Inception-Resnet V2 model is utilized. It's a pre-trained model, which has been trained on over 1.4 million photos and over 1,000 classes. Convolutional neural networks are used in the Inception-Resnet V2 model for image identification in order to extract features. In our study, 80% of the data were retained for training purposes, while 20% were kept for testing purposes. We retain the preceding layer's parameters in our transfer learning approach while removing the last layer, which was retained as testing data. In our transfer learning approach, we preserve the parameters from the preceding layer, discard the last Inception-Resnet V2 layer, and add a few more personalized layers. The final layer has the same number of output nodes as the number of categories in the dataset. We used a weight that has been previously trained on ImageNet. The completely modified layer was trained. 3,682,310 trainable parameters are available in total. The experiment uses the Adam optimizer with 256 batches, a learning rate of 1e-4, categorical cross-entropy as the loss function, and an input tensor of Inception-Resnet V2 set to (256, 256, 3). Rectified linear unit (Relu) was used as the activation function for the deep layer and Softmax for the output layer. Relu is linear for all positive values and 0 for all negative values. The model is quicker to train since the calculations are straightforward. Because it does not have the vanishing gradient problem that other activation functions have, this function is used. The number of output nodes in the top layer equals the number of dataset categories. After each layer in our extra layer—aside from the output layer—we utilized Dropout (0.5). For each hidden layer, there are 1080 neurons. And, final layer has in total 6 layers for 6 classification output (figure 3).

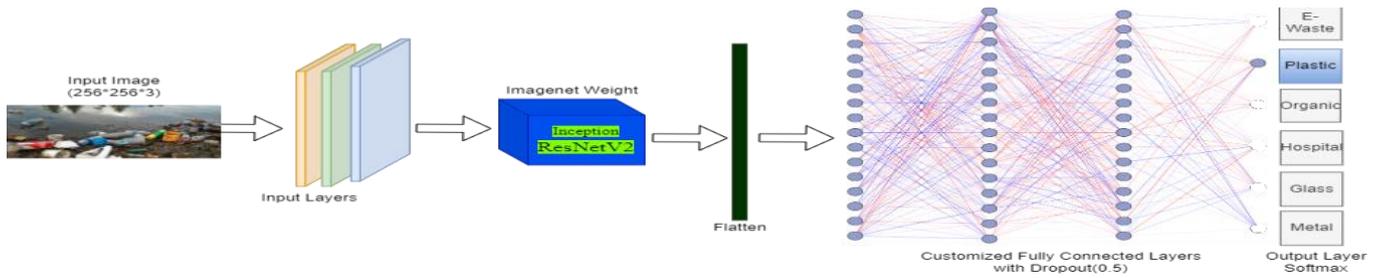

Figure 3: ConvoWaste Model with Dropout 0.5 and modified hidden layers

## IV. Experimental Setup for Object Detection

### A. Process of Creating Training Dataset

Initially, for six types of the waste materials, total 12000 images were captured in different lighting condition and orientations. To train up the network, 75 percent of the images were considered as the training dataset and rest of the 25 percent were used as the testing dataset. It is important to split the dataset into two sub-datasets as training and testing for eliminating overfitting and under-fitting of the custom model. In order to verify the precision of the custom dataset, the test dataset was used. The total number of pictures used for training and testing is given in table I.

TABLE I.    TOTAL QUANTITY OF PICTURES PER CATEGORY

| Categories | The total quantity of the pictures | | |
|---|---|---|---|
| | Training | Validation | Total |
| Plastic | 2000 | 400 | 2400 |
| Metal | 2000 | 400 | 2400 |
| Glass | 2000 | 400 | 2400 |
| Medical waste | 2000 | 400 | 2400 |
| Organic | 2000 | 400 | 2400 |
| E-waste | 2000 | 400 | 2400 |
| Total | 12000 | 2400 | 14400 |

We have collected the dataset by web scraping and Kaggle dataset site as there is no existing dataset for six different categories as a single.

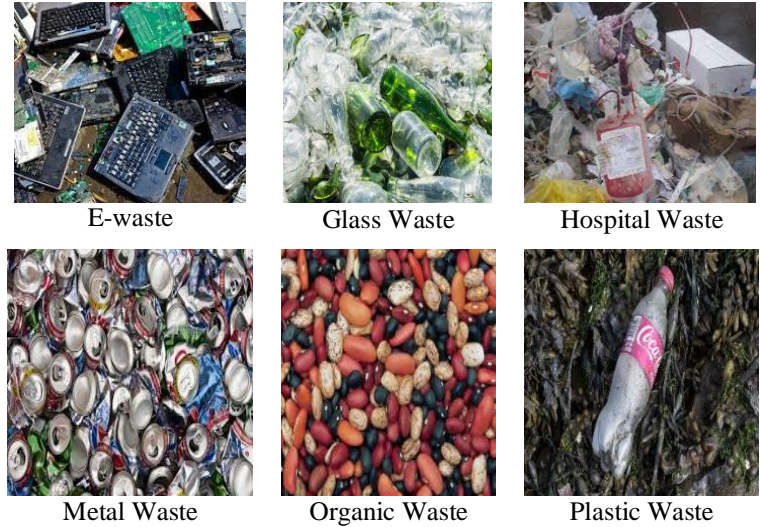

E-waste          Glass Waste          Hospital Waste

Metal Waste          Organic Waste          Plastic Waste

### B. Iteration for Achieving Higher Accuracy Level and Lower Loss level

The trained deep convolutional neural network requires several iterations to achieve its best possible output for image classification. The depicted graph in Fig. 4 shows the initial loss level of the network and the final accuracy level of the network after 100 epochs.

At the very beginning, the network shows the higher loss level. After doing several iterations, the trained network outperforms its initial level, which is around 90 percent and loss level of the trained network tends to decrease, which is around 10 percent.

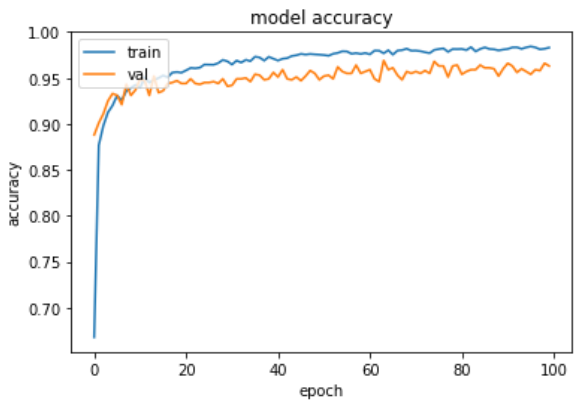

*Figure 4: Comparison of Model Training Accuracy & Validation Accuracy.*

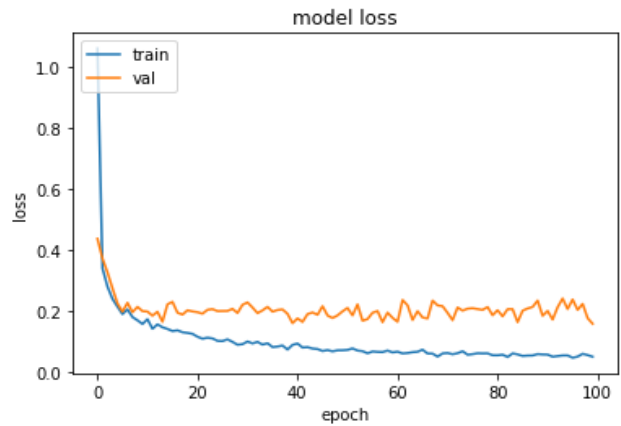

*Figure 5: Comparison of Model Training Loss & Validation Loss*

## V. Hardware Design and Implementation

Before implementing the project, physical design is important to be drafted first. Therefore, AutoCAD design is done before the implementation of the practical design. The total length of the system is taken as 12 feet and the width is 6 feet 5 inches to accommodate all the sensors, servomotors and camera module. The total length of the conveyor belt is taken as 15 feet to convey the waste from one end to another end by using the DC gear motor. The given design in Fig. 6 shows the complete setup of the system.

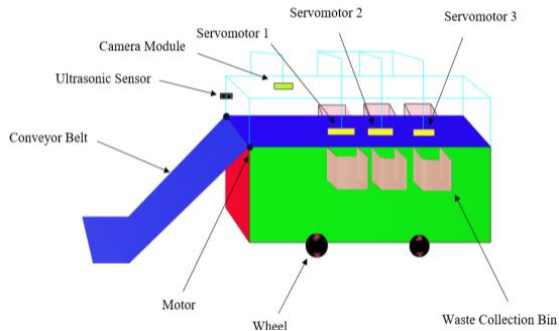

Fig. 6. *AutoCAD design of the system.*

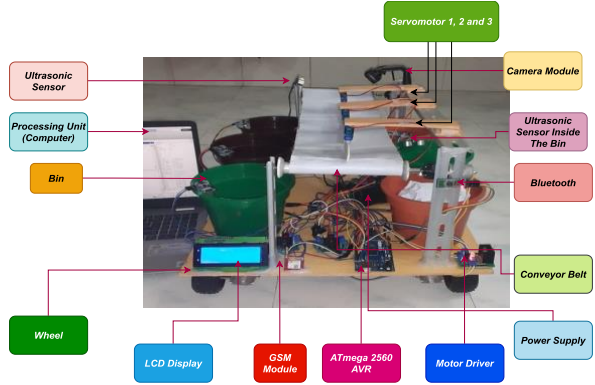

Fig. 7. *Overall hardware implementation of the system.*

According to the design of the system, a conveyor belt is made to convey the waste from one end to another end. Initially, it was planned to take as 15 feet but due to the budget limitation, it has been structured as 10 feet using DC gear motor. Afterwards, a physical infrastructure is made to hold the camera module, ultrasonic sensor and all the servomotors. There are six types of waste collection bins to collect different types of classified waste. In addition, a GSM-based communication medium is used to monitor the waste level inside the bins and the time to trash out the bins filled with garbage. Finally, this device has been made in such a way so that it can aid the dumping process remotely via an android controlled app based on Bluetooth communication. The inserted picture in Fig. 7 demonstrates the overall hardware implementation of the system.

## VI. Result Analysis

The overall detection and separation of the system are quite satisfactory. When an object is detected, a signal is sent to the servomotor controller from the processing unit (Computer) to separate the detected waste inside the corresponding bin. Afterwards, the cantilever of the servomotor comes to its initial position to be ready for the next detected waste. The depicted pictures in Fig. 8 (a) and (b) show the proper placement of the servomotor's cantilever.

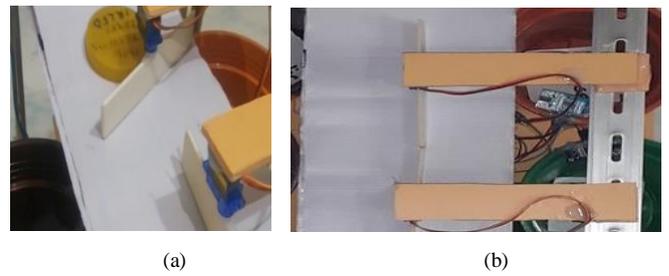

Fig. 8 (a) and (b). *Proper placement of the servomotor's cantilever.*

The display module of the system extracts the final output of the network and it shows the type of the waste properly. In addition, it counts the bin number when the waste is placed inside. This process helps the system to trash out the exact bin filled with garbage. The inserted picture in Fig. 8 (a) and (b) display the final output of the system.

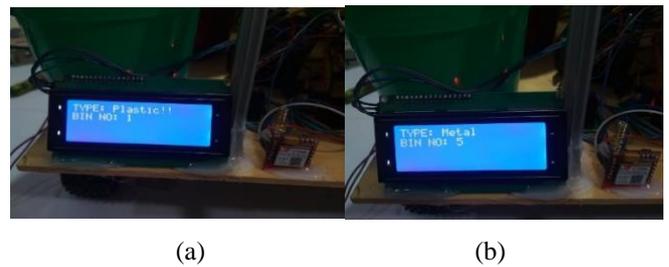

Fig. 9 (a) and (b). *Integration of display module with the system.*

Table II shows the time taken for detection and accuracy of the network. The above table II describes that the maximum accuracy level of the network is 96 percent that is for organic waste and the minimum accuracy level of the network is 88 percent that is for e-waste. To detect the waste, minimum time is taken for plastic, which is around 3 seconds and maximum time is taken for e-waste that is around 9 seconds.

TABLE II.     Time Taken for Detection And Accuracy

| Categories | Test Images | Classified | Accuracy (%) | Time (Sec) |
|---|---|---|---|---|
| Plastic | 50 | 47 | 94 | 3 |
| Metal | 50 | 47 | 94 | 5 |
| Glass | 50 | 45 | 90 | 6 |
| Organic | 50 | 48 | 96 | 6 |
| Medical waste | 50 | 46 | 92 | 4 |
| E-waste | 50 | 44 | 88 | 9 |

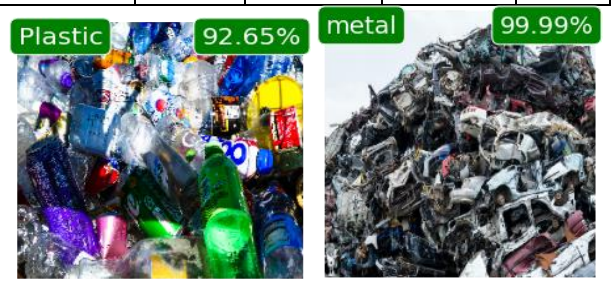

*Figure 10: Real world image test result after the train model.*

Table III: Comparison of different research work.

| Research Work | Dataset | Accuracy |
|---|---|---|
| DeepWaste [26] | DeepWaste database | 88% |
| Intelligent Waste Classification [27] | Trash image dataset | 87% |
| Recyclable Waste Classification [29] | Five Categories Waste Image | 88% |
| **Ours Research work** | **Six Categories Waste Dataset** | **98.30%** |

According to table III, our ConvoWaste model outperformed all prior state-of-the-art findings for several categories. In Figure 10, we have tested our model with an unseen dataset from the real world, and it accurately predicts the expected output with a high level of confidence.

## VII. CONCLUSION

All over the world, the governments are anxious about the sustainability of the country and circular economy (CE) to optimize the resource utilization. The focus on sustainability and CE, this trend will continue at least until 2050 as suggested by United Nations (UN) and others developed countries. This project is a small step towards fulfilling these two goals through the proper segregation of waste based on deep learning and image processing technique. A prototype has been implemented and tested successfully based on the proposed design in a lab environment. The results of this system showed higher accuracy level underlying deep learning algorithm. In this paper, for the training purpose, 12000 images were considered to train up the network but for the large-scale application. We have achieved the maximum accuracy 98% of the ConvoWaste.